\documentclass[review]{elsarticle}

\journal{Pattern Recognition}

\usepackage{graphicx}
\usepackage{comment}
\usepackage{amsmath,amssymb} 
\usepackage{color}
\usepackage{epsfig}
\usepackage{xcolor}
\usepackage{subfigure}
\usepackage{enumerate}
\usepackage{algorithm}  
\usepackage{algorithmic}
\usepackage{multirow}
\usepackage{array}
\usepackage{enumitem}
\usepackage{helvet}  
\usepackage{courier}  
\usepackage{url}

\usepackage{multirow}
\usepackage{booktabs}
\usepackage{tabularx}
\usepackage{bbding}
\usepackage{float}
\usepackage{setspace}
\usepackage{bbm}

\usepackage{caption}
\usepackage{color}
\usepackage{nth}
\usepackage{xspace}
\usepackage{stackengine}
\usepackage{placeins}

 \usepackage{xspace}

\newcommand{\RV}[1]{{{#1}}}

\makeatletter
\DeclareRobustCommand\onedot{\futurelet\@let@token\@onedot}
\def\@onedot{\ifx\@let@token.\else.\null\fi\xspace}

\def\eg{\emph{e.g}\onedot} 
\def\ie{\emph{i.e}\onedot} 
 
 \def\vs{\emph{vs}\onedot}

\makeatother

\begin{document}

\begin{frontmatter}

\title{Self-Regularized Prototypical Network for Few-Shot Semantic Segmentation}



\author[mymainaddress]{Henghui Ding}
\ead{henghuiding@gmail.com}
\cortext[mycorrespondingauthor]{Corresponding author}

\author[mymainaddress]{Hui Zhang}

\author[mymainaddress]{Xudong Jiang\corref{mycorrespondingauthor}}

\address[mymainaddress]{School of Electrical and Electronic Engineering, Nanyang Technological University, Singapore, 639798}
\begin{abstract}
The deep CNNs in image semantic segmentation typically require a large number of densely-annotated images for training and have difficulties in generalizing to unseen object categories. Therefore, few-shot segmentation has been developed to perform segmentation with just a few annotated examples. In this work, we tackle the few-shot segmentation using a self-regularized prototypical network (SRPNet) based on prototype extraction for better utilization of the support information. The proposed SRPNet extracts class-specific prototype representations from support images and generates segmentation masks for query images by a distance metric - the fidelity. A direct yet effective prototype regularization on support set is proposed in SRPNet, in which the generated prototypes are evaluated and regularized on the support set itself. The extent to which the generated prototypes restore the support mask imposes an upper limit on performance. The performance on the query set should never exceed the upper limit no matter how complete the knowledge is generalized from support set to query set. With the specific prototype regularization, SRPNet fully exploits knowledge from the support and offers high-quality prototypes that are representative for each semantic class and meanwhile discriminative for different classes. The query performance is further improved by an iterative query inference (IQI) module that combines a set of regularized prototypes. Our proposed SRPNet achieves new state-of-art performance on 1-shot and 5-shot segmentation benchmarks.
\end{abstract}

\begin{keyword}
Few-shot segmentation, prototype, prototypical network, self-regularized, non-parametric distance fidelity, iterative query inference, SRPNet, CNN.
\end{keyword}

\end{frontmatter}

\section{Introduction}
Deep learning has achieved enormous success in segmentation~\cite{FCN, ding2021vision, gu2018recent,VLT,liu2021towards,liu2022instance,ding2022deep}, thanks to large-scale segmentation datasets and powerful convolutional neural network (CNN)-based architectures~\cite{liu2019feature,chiou2021recovering,li2021mine,mei2019deepdeblur,li2021else,cai2021unified,wang2022learning}. However, it is expensive and laborious to obtain large datasets with pixel-wise annotations, which makes it impractical when dealing with real-world problems with numerous categories~\cite{zhang2021prototypical,ding2020phraseclick,sun2021m2iosr}. 
In addition to the difficulty of getting finely-labeled training data, poor generalization capability is witnessed when transferring the knowledge learned from training data to unseen classes, or when the training and testing images vary significantly in appearance. 
To address the aforementioned challenges, a rising task, the few-shot segmentation is proposed~\cite{OSLSM, SG-one, bhunia2019deep,liu2021few}. The few-shot segmentation is defined as: giving a small set of examples termed support images and their annotation masks, to segment unseen images termed query images based on the feature extraction of the support images. 
Specifically, in one-shot segmentation, each query image has reference to only one support image.

\begin{figure}[t]
\begin{center}
  \includegraphics[width =0.96\textwidth]{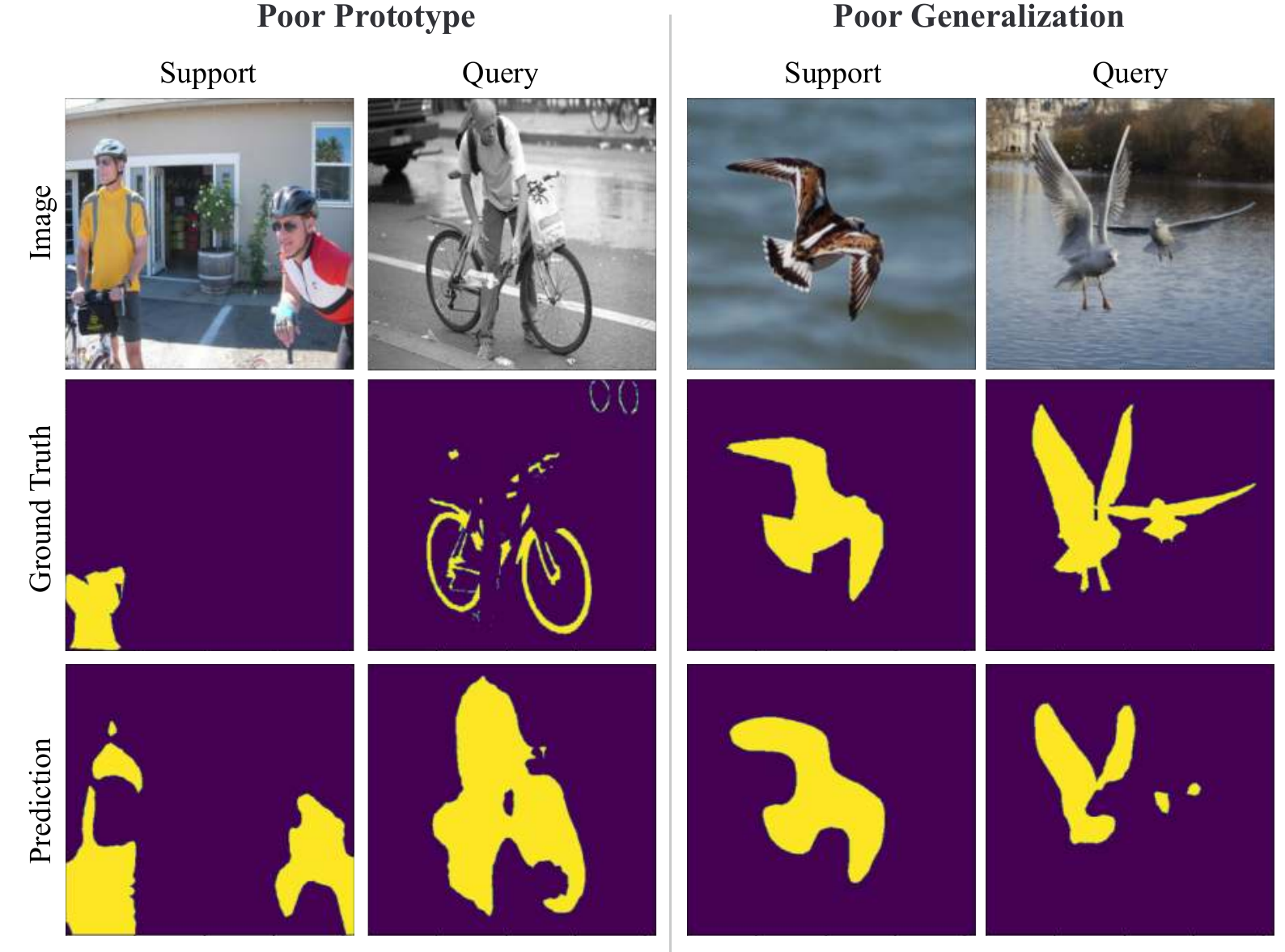}
\end{center}
\vspace{-0.36cm}
\caption{Two main problems in few-shot segmentation, namely, poor prototype and poor generalization. The poor prototype means that the generated support prototypes are low-quality and even cannot restore the segmentation mask of the support image. The poor generalization means the network cannot achieve satisfying predictions of query mask by the support prototypes.}
\vspace{0.2cm}
\label{fig:1}
\end{figure}

Existing few-shot segmentation approaches can be divided into either prototypical methods~\cite{dong2018few, PANet, FWP} or parametric methods~\cite{yan2019dual, siam2019amp, fan2020fgn}, based on how they establish linkages between support set and query set. In prototypical methods, the query mask is obtained by calculating the similarity between query features and the prototypes extracted from the support features~{by masked average pooling, where a prototype is a feature vector that contains class-related representations}. 
Whereas, in parametric methods, the knowledge extraction and query segmentation are combined together, \eg, by applying convolution over the concatenation of query features and support features. In this work, the prototypical manner is adopted for its characteristics of lightweight, compact, and robust to over-fitting. Moreover, based on the architecture of CNN, we propose a self-regularized prototype network (SRPNet) that enhances the few-shot segmentation through supervised prototype generation, improved similarity measuring, and iteratively optimized segmentation.

{We analyze the occurrence of unsuccessful mask predictions and attribute them to two challenges: 1) poor prototype that happens when the prototypes generated by masked average pooling over support features are not representative and can hardly restore even the mask of support set itself, and 2) poor generalization, which is usually induced by less-effective similarity measures, diverse appearance of objects in support set and query set, and the inherent difficulty of transferring learned knowledge to unknown classes.} Examples of the both challenges are collected from experiments on PASCAL-5$^i$ and shown in \figurename~\ref{fig:1}. Despite extensive efforts devoted to improving poor generalization~\cite{CANet, FWP}, the poor prototype remains an overlooked problem with plenty of room for improvement. We, however, consider the prototype generation a process most worth exploring and optimizing, since if the prototype obtained from the support feature is not an apposite representative, it can hardly achieve satisfying performance on the query set. 
Here, we evaluate the generated prototypes by applying them reversely onto the support set itself and try to restore the support masks using the same prediction method as for the query images. In fact, we do have observed in experiments that the generated prototypes cannot restore the support mask well - losing many details and sometimes even not consistent (see the first column in \figurename~\ref{fig:1}). Such results indicate that the generated prototypes lack discriminative representative and explicit supervision on the prototype generation is required. 
Prototype extraction (by masked average pooling) is a relatively fixed process and heavily depends on feature extraction. Without effective supervision during the end-to-end training, it always happens that the embedded feature map cannot deliver a prototype that comprehensively outlines the object. Some previous works share a similar spirit and introduce supervision onto the interaction of support and query features, concretely through the prototype alignment regularization~\cite{PANet} or the cross-reference module~\cite{CRNet}. However, they neither directly evaluate the quality of generated prototypes nor give explicit supervision on the prototype generation process. 

Here, we propose a more direct yet effective supervision module termed self-regularized prototype~(SRP) generation to evaluate and enhance the prototypes. 
Firstly, we generate prototypes by masked-average-pooling of support masks over embedded support features. The generated prototype is then applied back to the support feature to restore the support mask. 
{The quality of the generated prototype is measured by its predictive accuracy on the support image, \ie, how well the prototype can restore the support mask.}
We raise that the quality of a prototype can be evaluated by the extent to which it restores the support mask. Thus, we apply the prototype back to the support feature to perform segmentation.
The ``extent'' is quantified through certain evaluation metrics (\ie, mIoU and Binary-IoU). 
Poor prototype examples in \figurename~\ref{fig:1}) indicate that the prototypes cannot well recover support mask due to a lack of distinction and representativeness. Explicit supervision is desired for prototype generation.
To facilitate the network to learn better prototypes, we introduce an additional loss over the restored support mask, and support mask ground truth is used 
for regularizing the prototype generation in reverse. The model is motivated to produce prototypes with enhanced consistency and comprehensiveness for support and query sets, offering improved segmentation performance. The proposed regularization is only imposed on the support features, avoiding repeated interaction between the support and query set. The computational cost for regularization occurs only in training, leaving inference free. Once the feature maps are extracted, no extra learnable parameters are introduced and thus it is less prone to over-fitting.
To retain more details, we adopt a pyramid structure, where feature maps at various levels of the backbone extractor are concatenated and down-sampled to proper channel size, forming a final feature map with sufficient global scenery representation.

{Although our prototypes are well learned, plain generalization will make such improvement less pronounced on query set. For solving this, we adopt an iterative query inference~(IQI) module where a collection of prototypes are used for segmentation. Each prototype included is modified based on the initial prototype learned by SRP. Besides, we adopt a new metric~-~the fidelity~-~for measuring the similarities. Both fidelity and cosine similarity measure the angle rather than the absolute distance between two separate vectors. However, unlike cosine similarity that has a symmetrical value space about the origin while the negative part is meaningless in image segmentation, our fidelity distributes only between 0 and 1 and exhibits more evident distinctions for vectors in different directions.}

In short, the main contributions of our work are:
\begin{itemize}
\setlength\itemsep{1em}
  \item We propose a direct yet effective self-regularization module. Prototypes are generated, evaluated, and regularized under the supervision of support masks, which differs from existing works.
  \item We adopt fidelity as the distance metric in prototype generation for the first time, which provides a more evident distinction for vectors.
  \item We adopt an iterative query inference module, which uses a collection of prototypes for segmentation and improves the generalization ability for query inference.
  \item We achieve new state-of-the-art performance on two few-shot segmentation benchmarks.
\end{itemize}

\section{Related Work}
\subsection{Semantic Segmentation}
Semantic segmentation is a task that labels each pixel in an image by the most appropriate semantic category from the predefined ones~\cite{liang2022expediting,ding2019semantic,ding2021interaction, ding2019boundary}. Most recent methods since the first Fully Convolutional Network (FCN) by Long et al.~\cite{FCN} are constructed by deep convolutional neural networks and achieved remarkable improvements in segmentation performance. However, FCN lacks strategies to dig global context and spatial details. Dilated convolutions~\cite{DeepLabv2, fu2020contextual, ding2018context} are advantageous in remaining spatial resolution while enlarging the receptive fields. In addition, pyramid structure, e.g., feature pyramid network~\cite{ding2020semantic,wang2019bi, shuai2018toward, wang2021knowledge,wang2019dermoscopic} or pyramid spatial pooling~\cite{PSPNet}, is widely employed to capture multi-scale features. In this work, we perform dense prediction following the FCN. Dilated convolutions and pyramid structure are adopted to improve the feature extraction capabilities.

\subsection{Few-shot Classification}
The target of few-shot learning is to capture new concepts by just a few examples~\cite{zhang2021meta}. Existing work can be divided into three mainstreams, which are based respectively on metric learning~\cite{snell2017prototypical, vinyals2016matching}, optimization learning~\cite{koch2015siamese, ravi2016optimization}, and graph neural network~\cite{garcia2017few, liu2018learning}. It was first proposed by Oriol et al.~\cite{vinyals2016matching} to encode the input into embedded features and perform weighted nearest neighbor matching for classification. Snell et al.~\cite{snell2017prototypical} proposed a prototypical network that uses a feature vector called prototype to represent each class. Recently, the concept of few-shot has been extended from classification to the challenging semantic segmentation, aiming to segment objects of unseen categories with limited annotated images. Our work applies the prototypical network to few-shot segmentation, in which features at each pixel is classified by its distance to class prototypes.

\subsection{Few-shot Segmentation}
As an extension of few-shot classification, the few-shot segmentation aims to generalize the segmentation capability to new categories, while being supported by only a few annotated examples~\cite{liuTMM22}. As the first to apply the spirit of few-shot classification to segmentation tasks, Shaban et al.~\cite{OSLSM} proposed a two-branch architecture, where a conditioning branch generates free parameters that are used to modulate the FCN-based segmentation branch. Such parametric modules fuse the features extracted from the support set for the execution of segmentation. Similarly, Rakelly et al.~\cite{co-FCN} realized segmentation by decoding the concatenation of support and query features.

Prototype methods are popular in few-shot segmentation. For example, Dong et al.~\cite{dong2018few} proposed a dense prototype learning and adopted Euclidean distance as the metric. 
Wang et al.~\cite{PANet} adopted a simpler design, in which prototypes are incorporated with query features to generate annotation masks directly. 
Recently, PPNet~\cite{liu2020part} and PMMs~\cite{yang2020prototype}, which are also a prototype networks, learns to use the distance to each class prototype to perform metric space classification. In PPNet, they proposed part-aware prototypes to highlight the fine-grained features, and try to enhance the prototypes based on both labeled and unlabeled images. In PMMs, the prototype represents a part of an object instead of a class of samples, and the prototype network does not involve a single sample or a mixed prototype of a class of samples. PFENet~\cite{tian2020pfenet} generates training-free prior masks, and enriches query features with support features and the prior masks.

In most prototypical networks~\cite{PANet,liu2019prototype,liu2020part}, cosine similarity is used to weight features for the foreground predictions. In our paper, a new distance metric, fidelity is used for generating guidance map. The fidelity measures the distance between the respective density matrices of the query feature and the prototype and has a more evident distinction for vectors in different directions, as well as improved segmentation performance. 

The interaction between query and support branches has drawn attention recently. For example, Wang et al.~\cite{PANet} proposed a reverse segmentation that uses the query image with its predicted annotated mask to segment the original support images.
Their routine was to perform the few-shot segmentation in a reverse direction, using the query image with its predicted mask to segment the original support images. In this way, they claimed to be able to generate more consistent prototypes between support and query, offering better generalization performance. 
A similar method is seen in Liu et al.~\cite{CRNet}, where the query set and the support set are concatenated to produce foreground predictions for the purpose of training supervision. In our work, we propose a most direct yet effective prototype regularization. Concretely, the generated prototypes are applied reversely on the support features to restore the support masks. We ensure that the prototypes best recovers support features as a prior of its generalization to query features. The segmentation masks of both the support and query sets are adopted as training supervisions.

\section{Method}

\subsection{Problem Description}
{Few-shot segmentation~\cite{OSLSM,PANet,tian2020pfenet} aims at segmenting objects based on the support information from just a few annotated training images. Each few-shot segmentation task $\mathcal{T}$ (also named as an episode $\mathcal{T}$) consists of a support set $S$ supplied with ground-truth masks and a query set $\mathcal{Q}$.} The support set $\mathcal{S}=\{I, M\}$ contains only a few image - mask pairs. The query set $Q$ contains $N_q$ image-mask pairs. {Specifically in a $C$-way $K$-shot segmentation task where there are $C$ classes and $K$ support samples for each of the $C$ classes, we denote the support set as $\mathcal{S}=\{(I_{c,k},M_{c,k})\}$, where $k\in\{1,\cdots,K\}$ and $c\in\{1,\cdots,C\}$, indicating that $K$ image-mask pairs per semantic class from the $C$ classes are included. For example, for the class $c$, where $c\in\{1,...,c,...,C\}$, there are $K$ image-mask pairs as support samples, \ie, $\{(I_{c,1},M_{c,1}),(I_{c,2},M_{c,2})...,(I_{c,K},M_{c,K})\}$. The semantic classes are consistent in the support set and the query set, \ie, the query images are segmented to the $C$ classes (and the background class) provided in the support set. The goal is summarized as, to generate a model that, when given a support set $S$, predicts the annotation masks for the query images $Q$.}

The overall classes are divided into two non-overlapping sets of classes $C_{seen}$ and $C_{unseen}$, used for training and testing respectively. The training set $D_{train}$ is constructed from images of classes $C_{seen}$ and the testing set $D_{test}$ is constructed from images of classes $C_{unseen}$. Both the training set $D_{train}$ and the testing set $D_{test}$ contains lots of episodes $\mathcal{T}$. {The training set $D_{train}$ is composed of all image-mask pairs that contain at least one pixel in the segmentation mask from training classes. The testing set is randomly sampled. The testing classes $C_{unseen}$ are different from the training classes $C_{seen}$.}
Notably, for images in $D_{train}$, annotations of $C_{unseen}$ objects are excluded from the $D_{train}$ by labeling them as background, while the images are included as long as there is an object from $C_{seen}$ present.
The features are embedded from images by initializing CNN with pre-trained weights. During training, the model gains knowledge from the support set and then applies the knowledge to segment the query set. As each episode contains different semantic classes, the model can be well generalized after training. After obtaining the segmentation model $L$ from the training set $D_{train}$, we evaluate its few-shot segmentation performance on the test set $D_{test}$.

\begin{figure*}[t]
\begin{center}
  \includegraphics[width =1\textwidth]{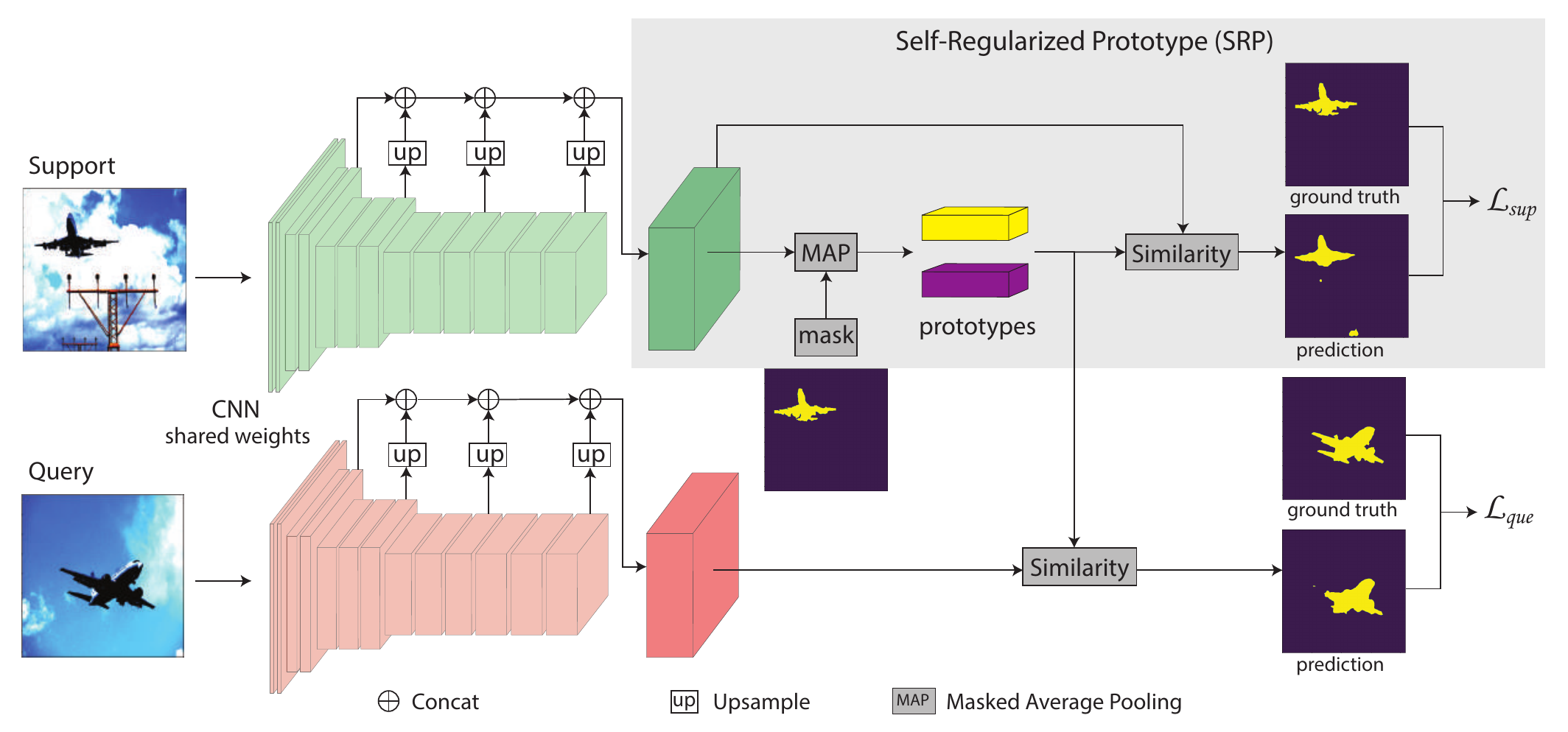}
\caption{Architecture overview of our training model {in a 1-way 1-shot example}. We embed the support and query images into deep features by ResNet-50 with shared weights. Features from multiple levels are up-sampled and concatenated to form the final feature map. The prototypes, {including foreground one in yellow and background one in purple, are generated by masked average pooling over the support features.} Both the support images and the query images are segmented by computing the pixel-wise similarity between the prototypes and feature maps. The support loss $\mathcal{L}_{sup}$ and query loss $\mathcal{L}_{que}$ are used for end-to-end training.}
\vspace{0.16cm}
\label{fig:2}
\end{center}
\end{figure*}

\subsection{Architecture Overview}
As opposed to the parametric way that fuses the support and query features to generate segmentation masks, we aim to learn an accurate and robust class-specifical prototype representation. The learned prototype is good enough to represent the corresponding class and can deal with different kinds of objects/images, which performs segmentation via non-parametric metric learning. The overall architecture of the proposed training model is shown in \figurename~\ref{fig:2}. During each episode, images from the support and query set are embedded indiscriminately into deep features by a shared backbone. The masked average pooling (MAP) is then applied over the support features to produce prototypes. MAP is also useful in filtering background noise. The query images are segmented by assigning the class with the most similar prototype to each pixel. A self-regularization is applied to the prototypes, by segmenting the support images using their own prototypes. This operation encourages the few-shot model to generate more accurate prototypes as a prior to improving the consistency between support and query set.

Following~\cite{CANet, tian2020pfenet}, we adopt ResNet-50~\cite{resnet} as the backbone network. 
Besides, prototypes generated by the high-level feature maps of support images are usually abstract. To retain more details, we adopt a pyramid structure, where feature maps at various levels of the backbone extractor are aggregated to provide sufficient global scenery representation.
A pyramid feature aggregation is applied to the convolutional blocks to generate feature maps at different semantic levels. As shown in \figurename~\ref{fig:2}, these pyramid features are up-sampled to the same size and their concatenations are processed by a $1\!\times\!1$ \texttt{Conv} to 512 channels.
No extra free parameters will be introduced then. We aim to optimize the above-mentioned backbone to learn a consistent embedding space and train the model end-to-end.

\subsection{Prototype Generation}
Our model learns representative and well-separated prototype representation, {a feature vector}, for each semantic class, including the background, based on the prototypical network. We produce the foreground objects and background prototypes by applying the support masks over the feature maps separately. Given a support set $S=\{(I_{c,k},M_{c,k})\}$, where $c$, $k$ indexes the class and the shot respectively, let $F_{c,k}$ be the support feature map from the backbone and corresponds to the image $I_{c,k}$, the prototype of class $c$ is computed by masked average pooling:
\begin{equation}
    p_c=\frac{1}{K}\sum_k\frac{\sum_{x,y}F_{c,k}^{(x,y)}\mathbbm{1}(M_{c,k}^{(x,y)}=c)}{\sum_{x,y}\mathbbm{1}(M_{c,k}^{(x,y)}=c)},
\end{equation}
where $(x,y)$ indexes the coordinate location of pixels and $\mathbbm{1}(*)$ is an indicator function, outputting $0$ if $*$ is false or 1 otherwise. Similarly, we compute the background prototype by
\begin{equation}
    p_{bg}=\frac{1}{CK}\sum_{c,k}\frac{\sum_{x,y}F_{c,k}^{(x,y)}\mathbbm{1}(M_{c,k}^{(x,y)}=bg)}{\sum_{x,y}\mathbbm{1}(M_{c,k}^{(x,y)}=bg)},
\end{equation}
where $bg$ represents background. These prototypes are optimized non-parametrically during the end-to-end training as described in the following two sections.

\begin{figure}[t]
\begin{center}
  \includegraphics[width = 0.8\textwidth]{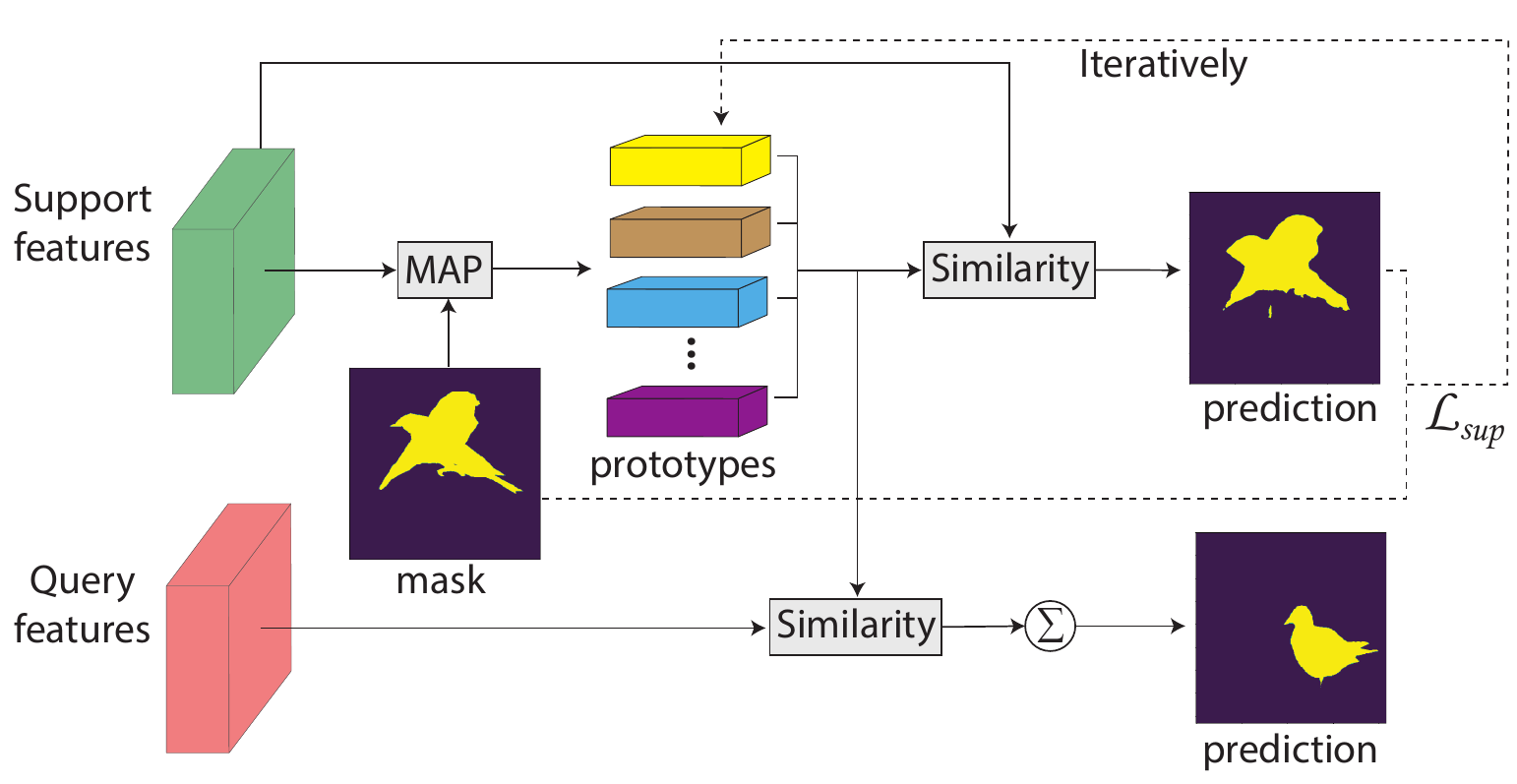}
\vspace{0.2cm}
\caption{Iterative Query Inference~(IQI). Multiple iteratively optimized prototypes, instead of a single prototype, are used for query inference. Each optimized prototype is based on the previous prototype’s evaluation of the supporting images. The final segmentation mask is generated by fusing the predicted annotations of each prototype.}
\vspace{0.12cm}
\label{fig:3}
\end{center}
\end{figure}

\subsection{Non-Parametric Distance Fidelity}
\label{section:34}
Here, we describe how the segmentation is performed using these prototypes and introduce our non-parametric distance fidelity.
We first calculate the similarity between each generated prototype over semantic classes (including background) and the feature vector $F_q^{(x,y)}$ at each position $(x,y)$ in the query feature. The softmax function is applied over the similarities to produce a probability map $\hat{M}_q$. Formally, given a distance function $\mathcal{D}$, let $\mathcal{P}=\{p_1,...,p_C\}\cup\{p_{bg}\}$ denote the prototype collection of prototype $p_c$ for class $c$ and $p_{bg}$ for background. For each $p_j\in \mathcal{P}$ (where $j=1,...,C\!+\!1$) and the query feature map $F_q$ we have
\begin{equation}
    \hat{M}^{(x,y)}_{q;j}=\frac{e^{\alpha\mathcal{D}(F_q^{(x,y)},p_j)}}{\sum_{j}e^{\alpha\mathcal{D}(F_q^{(x,y)},p_j)}},
\end{equation}
where $\alpha$ is an amplification factor.
The predicted segmentation mask is derived by 
\begin{equation}
    \hat{M}_q^{(x,y)}=\mathop{\arg\max}_{j} \hat{M}_{q;j}^{(x,y)}.
\end{equation}
The commonly adopted distance metric $\mathcal{D}$ is cosine distance~\cite{oreshkin2018tadam, PANet} and squared Euclidean distance~\cite{snell2017prototypical}. Here we introduce another distance metric, fidelity, which is defined as the distance between the respective density matrices of query features and prototypes. The density matrix as mentioned is defined formally as the outer product of a coordinate vector and its conjugate transpose. For each $F^{(x,y)}_q$ from query features and $p_j\in\mathcal{P}$ that both are reshaped to $1_{height}\times 1_{width}\times N_{channel}$, we first normalize them respectively by:
\begin{equation}
\begin{aligned}
    p_j&=\frac{p_j}{\lVert p_j\rVert},\\
    F^{(x,y)}_q&=\frac{F^{(x,y)}_q}{\lVert F^{(x,y)}_q\rVert}.
\end{aligned}
\end{equation}
The fidelity is then computed by:
\begin{equation}
    \mathcal{D}(F^{(x,y)}_q,p_j)= F^{(x,y)}_q\cdot (p_j^T \cdot p_j)\cdot (F^{(x,y)}_q)^T,
\end{equation}
where $p_j^T$ and $(F^{(x,y)}_q)^T$ is the transpose of $p_j$ and $F^{(x,y)}_q$, respectively. Both fidelity and cosine similarity measures distance from the direction rather than absolute value. Cosine similarity has a symmetrical value space about the origin, where the negative part is meaningless in the image segmentation. Unlike cosine similarity, the fidelity distributes only between 0 and 1 and thereby has a more evident distinction for vectors with different directions. 

Learning proceeds by minimizing the negative log-probability 
\begin{equation}
    \mathcal{L}_{que}=-\frac{1}{N}\sum_{x,y}\sum_{p_j\in\mathcal{P}}\mathbbm{1}(M_q^{(x,y)}=j)log\hat{M}^{(x,y)}_{q;j},
\end{equation}
where $N$ is the total number of spatial pixels, and $M_q$ is the ground truth mask of the query image.

\subsection{Self-Regularized Prototype (SRP)}

In most previous works, the training process is only driven by the loss between predicted masks and ground truth masks of query images. Attempts (\eg, prototype alignment~\cite{PANet}) were devoted to improving the generalization of knowledge from support images to query images, which encourages the consistency of mutual generalizability between support and query images. However, in addition to the prototype alignment, it is also worth verifying the quality of the prototypes generated by the support set, \ie, the extent to which these prototypes can restore the support masks. We consider the feature extraction a process worth exploring and optimizing since if the prototype obtained from non-parametric average pooling on the support features is not a good representative, there is no way it will get a good performance on the query set. In fact, we have observed in experiments that the generated prototype did not restore the support images well  - losing many details and sometimes even not consistent when using the same prediction method as for the query images (see \figurename~\ref{fig:1}). This requires supervision on the  prototype generation process from support features.

To address these issues, we propose a Self-Regularized Prototype (SRP) module that can not only evaluate but also improve the quality of prototypes by incorporating the loss between the prototype-restored support masks and the original support masks. 
In this module, we adopt a more direct and effective method, which is to apply the prototypes to the supporting features to form support score maps. 
When the score maps are applied to pixel-wise feature vectors of support images themselves, supporting masks will be generated. 
The additional loss of the prototype-restored support masks and ground truth support masks are used for regularizing the prototype generation reversely. The prototype generation is therefore encouraged to retain specific details and enhance the distinction between classes. Meanwhile, considering that the generalization on support images is undoubtedly the best, the evaluation of the generated prototypes on support images will impose an upper bound of segmentation. The performance on query set will never exceed this upper bound set regardless of generalization. The consistency and generality between support and query can also be reasonably quantified in this way.

The proposed SRP is illustrated in \figurename~\ref{fig:2}. The prototypes obtained from support set 
are employed to generate segmentation masks for not only query set but also support set, following the method explained in section~\ref{section:34}. For each $p_j\in \mathcal{P}$ and each position's feature in support set $F_s^{(x,y)}$, we have
\begin{equation}
    \hat{M}^{(x,y)}_{s;j}=\frac{e^{\alpha\mathcal{D}(F_s^{(x,y)},p_j)}}{\sum_j e^{\alpha\mathcal{D}(F_s^{(x,y)},p_j)}}.
\end{equation}
The predicted segmentation mask for support set is then given by 
\begin{equation}
    \hat{M}_s^{(x,y)}=\mathop{\arg\max}_{j} \hat{M}_{s;j}^{(x,y)}.
\end{equation}
Similarly the loss between support predictions and annotations
\begin{equation}
    \mathcal{L}_{sup}=-\frac{1}{N}\sum_{x,y}\sum_{j}\mathbbm{1}(M_s^{(x,y)}=j)log\hat{M}^{(x,y)}_{s;j}.
\end{equation}
The overall training loss is
\begin{equation}
    \mathcal{L}=w_s\mathcal{L}_{sup}+w_q\mathcal{L}_{que},
\end{equation}
where $w_s$ and $w_q$ are used to balance the contribution of support loss and query loss. By SRP, we provide a quantification of the goodness of the learned prototypes, and meanwhile impel the network to learn more accurate and comprehensive prototypes and to generate more consistent prototypes for support and query sets, offering better segmentation performance.

\subsection{Iterative Query Inference~(IQI)}
A single prototype, though fine-tuned by end-to-end training, is not sufficiently explicit to segment an object due to the limited details it can retain and the large variation in object appearance within the same category. Usually, the prediction merely indicates the rough position and shape of the objects. To solve this issue, we propose an iterative collection of prototypes in query inference to further enhance the representative and generalization of prototypes according to their evaluation on support set. It's not necessary to include iterative optimization of prototypes in the training, because few commonality is observed in the detailed expression among different objects. 
The structure of our Iterative Query Inference~(IQI) is shown in \figurename~\ref{fig:3}, in which the prototypes are collected iteratively by:
\begin{equation}
    p_{j;n} = p_{j;n-1}-\eta\partial\mathcal{L}_{sup}/\partial p_{j;n-1},
\end{equation}
where $\eta$ is the reference rate and $\partial\mathcal{L}_{sup}/\partial p_{j;n-1}$ is the gradients, $n\in\{1,\cdots,N\}$ and $N$ is the amount of collected prototypes. We fuse the predicted score maps $\{\hat{M}_{q;j}^n:n=1,\cdots,N\}$ to generate the final segmentation $\hat{M}_{q}$ by
\begin{equation}
    \hat{M}_{q}=\mathop{\arg\max}_{j} \sum_{n=1}^N\hat{M}_{q;j}^n\rho_n,
\end{equation}
where $\rho_n=IoU(\hat{M}_s,M_s)$ is the intersection-over-union metric of evaluating the support set.

Our few-shot segmentation model adopts prototype embedding. Predictions are performed on the computed feature maps, thereby requiring no extra passes through the network. In addition, the proposed regularization is only imposed on the support features, avoiding repeated interaction between the support and query set. The computational cost for regularization occurs only in training, leaving inference free. Once the feature maps are extracted, no extra learnable parameters is introduced and thus is less prone to over-fitting.
\section{Experiments}
\subsection{Dataset and Evaluation Metric}
We evaluate our approach on PASCAL-5$^i$~\cite{OSLSM} and MS COCO~\cite{MSCOCO}. PASCAL-5$^i$ is created from PASCAL VOC 2012~\cite{pascalvoc} by Shaban et al.~\cite{OSLSM}. The 20 categories included are divided evenly into 4 splits, each containing 5 categories. 
The training is executed in a cross-validation fashion, where 3 splits are used for training while the rest one is used for evaluation. Experimental results are reported on each of the 4 splits in testing. The categories in each split are as follows: split~1: $\{$bottle, boat, bird, bicycle, aeroplane$\}$; split~2: $\{$cow, chair, cat, car, bus$\}$; split~3: $\{$person, motorbike, horse, dog, dining-table$\}$; split~4: $\{$tv/monitor, train, sofa, sheep, potted-plant$\}$. {Categories are grouped according to the alphabetical order of category names so that the difficulty levels of different splits could be very large. Therefore, all methods have large performance difference among different splits.} The testing results are reported by the average of 5 independent runs, each containing 1000 episodes, and the random seed is different. Our models are also evaluated on MS COCO~\cite{MSCOCO}, with its 80 object classes evenly divided into 4 splits. $N_{query}=1$ is applied for all experiments.
The per-class Intersection over Union ($IoU$) is defined as $\frac{tp}{tp+fp+fn}$, where the $tp$, $fp$ and $fn$ is the count of true positives, false positives and false negatives, respectively. 
We use the $mean{\text -}IoU$ over all classes as the evaluation metric.

\begin{table}
\renewcommand\arraystretch{1.1} 
\small
\centering
\caption{Ablation studies on PASCAL-5$^i$ under 1-way 1-shot setting. $mean{\text -}IoU$ are shown for quantifying each contribution. (sup) denotes evaluation on support set.}
\begin{tabular}{c|ccccc}
\toprule
Methods & split-1 & split-2 & split-3 & split-4 & mean\\ 
\hline
Cos & 54.2 & 60.5 & 46.5 & 53.3 & 53.6 \\
F & 56.1 & 62.2 & 48.7 & 54.6 & 55.4 \\
F~+~SRP & 60.2 & 67.2 & 54.6 & 56.7 & 59.7 \\
F~+~IQI & 58.7 & 66.1 & 51.9 & 55.4 & 58.0\\
F~+~SRP~+~IQI & 62.8 & 69.3 & 55.8 & 58.1 & 61.5\\
\hline
F (sup) &71.6  &78.2  &74.4  &66.6  &72.7  \\
F+SRP (sup) &78.4  &86.1  &81.3  &74.9 &80.2 \\
\bottomrule
\end{tabular}
\label{tab:1}
\end{table}

\subsection{Implementation Details}
The ResNet-50~\cite{resnet} is adopted as our feature extractor, and shared weights are applied to support and query sets. We reserve the first 5 convolutional blocks, while discarding other layers. Input images are resized to (417,~417) and augmented using random horizontal flipping. You may notice later in qualitative results that the details of some objects in the support mask are too fine to be recognized due to the resizing process, but our predictions are not affected. The batch size is 1. 
To maintain large spatial resolution, the last two down-sampling operations are discarded, \ie, the output stride is 8.
Following~\cite{DeepLabv2, PSPNet}, dilated convolutions are used in the last two blocks to enlarge the receptive fields.
The model is trained end-to-end by SGD for 30,000 iterations. The momentum is set to 0.9, the learning rate is initialized to 0.001 and decreased by 0.1 every 10,000 iterations, and the weight decay is set to 0.0005.

\subsection{Ablation Studies}
Ablation studies are conducted to verify the effect of each of our contributions on performance. The experiments in \tablename~\ref{tab:1} include: \textbf{Cos} - a baseline prototype network with cosine similarity; \textbf{F} - a baseline prototype network with fidelity as the distance metric; \textbf{F+SRP} - extends the baseline model with the prototype regularization on support set; \textbf{F+IQI} - extends the baseline model with iterative query inference; \textbf{F+SRP+IQI} - represents our full approach. {Our baseline is \textbf{Cos} in \tablename~\ref{tab:1}, \ie, ResNet-50 based prototype network~\cite{snell2017prototypical} with cosine similarity. We choose this one because cosine similarity is commonly used in prototypical methods~\cite{snell2017prototypical, PANet} and we aim to enhance the prototypical way by analyzing the weaknesses of the prototypical methods and proposing SRPNet to address these weaknesses.} We also apply the F+SRP on support set in each split. The inference of supporting images is only related to the quality of the prototype generation and has nothing to do with the generalizability of the model. Meanwhile, the improvement will be less pronounced due to the generalization from support set to query set  - which is consistent with our results when we compute and compare the difference between mean-IoU of F+SRP and F, on support set and query set respectively. The IQI is applied for improving the generalizability and mitigating such performance loss, given that the prototypes are trained sufficiently discriminative for classes. In other words, the improvements observed on query set come from two aspects, one is the optimized prototype extraction, and the other is the knowledge combination from versatile prototypes during inference. Specifically, the number of iteratively optimized prototypes in inference is set to 5.

Table~\ref{tab:1} shows the ablations with the aforementioned variants in the one-shot setting on PASCAL-5$^i$. 
The ablation studies clearly show the contributors to our performance improvement. The performance gain brought by our F (fidelity) \vs cosine distance metric is 1.8\%. Compared to F, our most important contribution of SRP improves the $mean{\text -}IoU$ by 4.3\%. However, the upper limit of the SRP contribution can be as high as 7.5\%, when applying the same prototypes on support set itself. F+IQI imposes an improvement of 2.6\% compared to F. Combining SRP and IQI improves the overall performance by 6.1\%, which is higher than applying either of SRP and IQI separately, but slightly lower than the sum of their independent improvements (which is $\sim$6.9\%). While SRP optimizes the prototype generation (revealed by that it adapts much better to support set), the IQI effectively generalizes the learned prototypes to new classes in testing.

\begin{figure}[hbtp]
\begin{center}
\includegraphics[width =\textwidth]{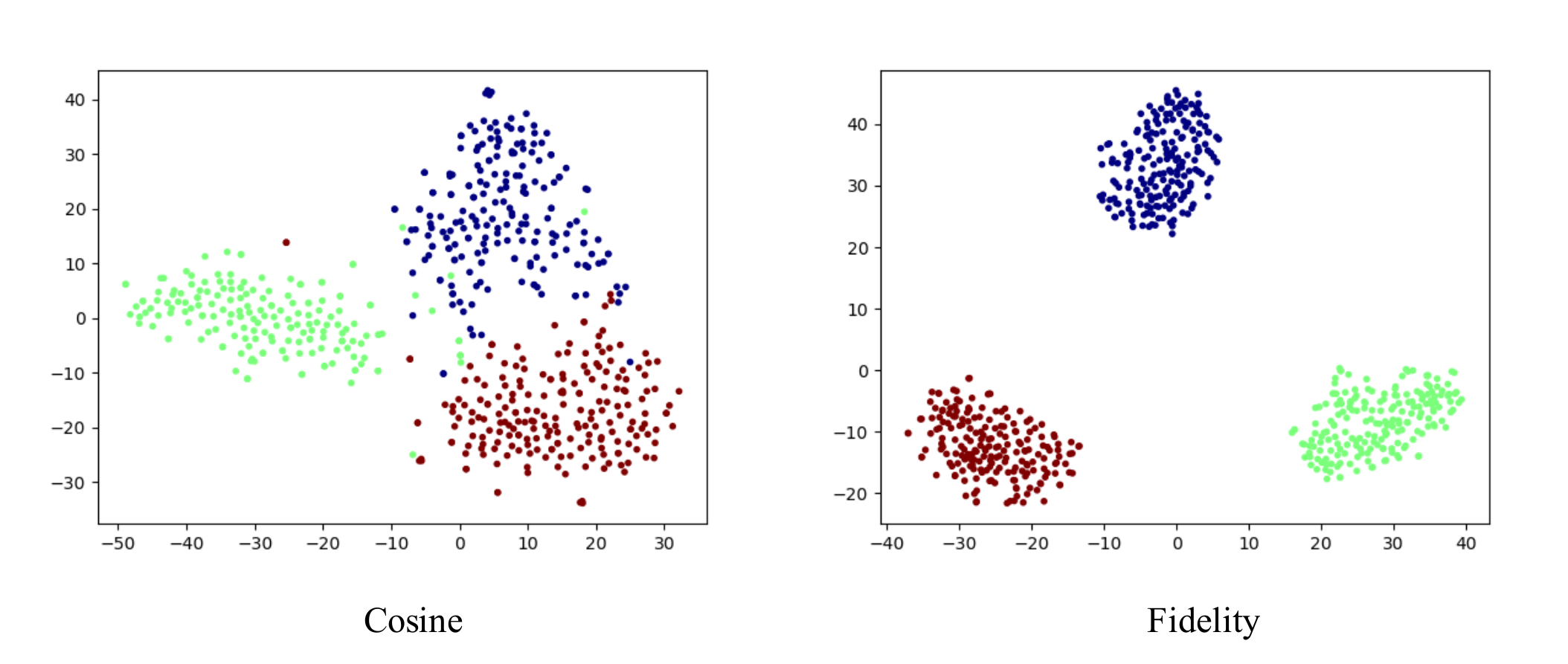}
\end{center}
\vspace{-0.36cm}
\caption{{tSNE visualization of support features by Cosine and Fidelity. Each color represents a class and each point represents a feature vector. Compared to cosine distance, the fidelity distance has a more distinct inter-class feature distribution (\eg, green \vs blue) and a more compact intra-class feature distribution (\eg, green), which supports the superior of fidelity over cosine distance.}}
\label{fig:tSNE}
\end{figure}
{In \figureautorefname~\ref{fig:tSNE}, we visualize the tSNE of features by cosine distance (left) and fidelity distance (right), respectively. Both fidelity and cosine similarity measure differences in angles rather than absolute distances. Cosine similarity has a symmetrical value space with respect to the origin, where the negative part has no meaning in image segmentation. The value space for fidelity is positive. Compared to cosine similarity, there is a more pronounced difference in the fidelity of vectors in different directions. As shown in \figureautorefname~\ref{fig:tSNE}, compared to cosine distance, the fidelity distance has a more distinct inter-class feature distribution (\eg, green \vs blue) and a more compact intra-class feature distribution (\eg, green), which supports the superior of fidelity over cosine distance.}

\begin{figure*}[t]
\begin{center}
\includegraphics[width =\textwidth]{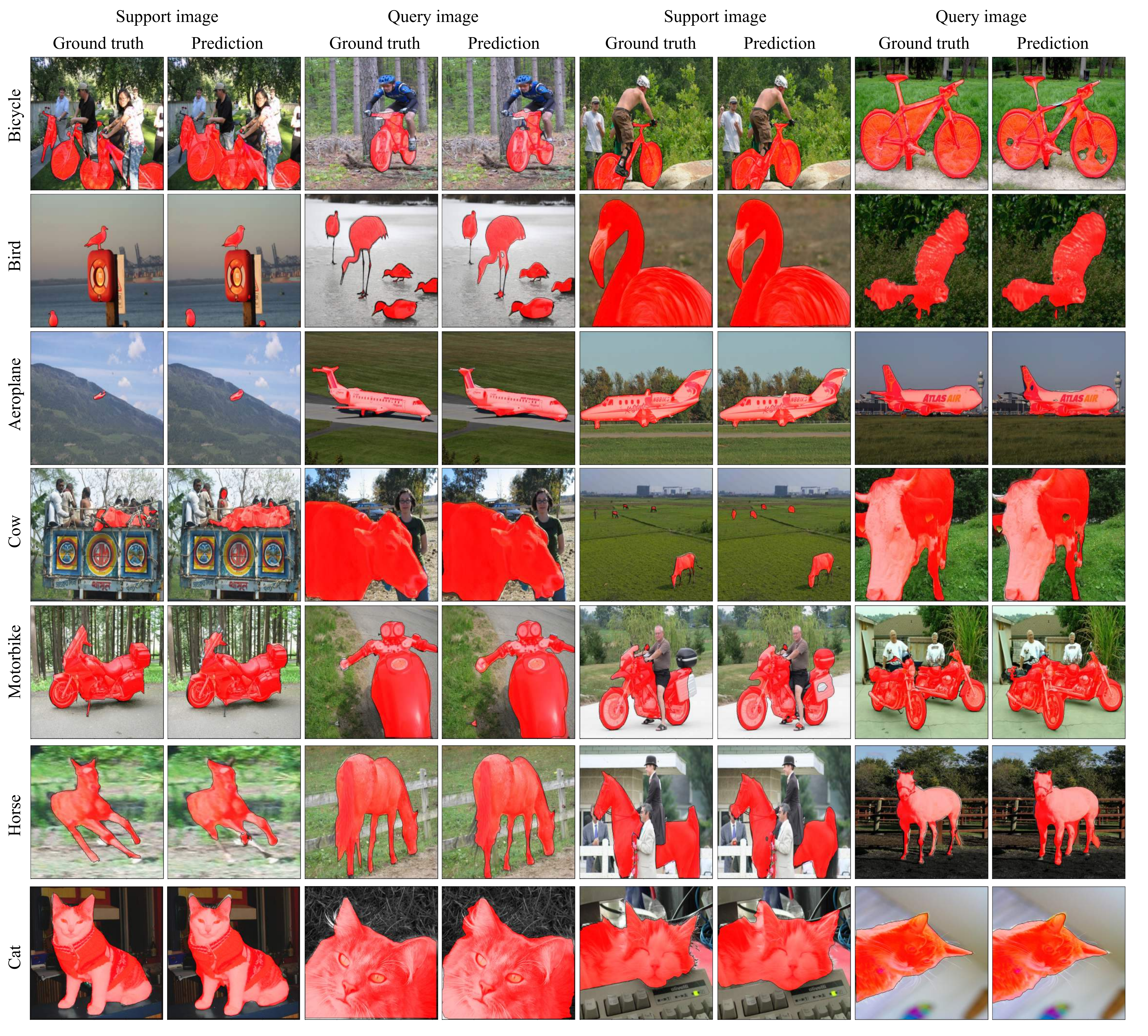}
\end{center}
\vspace{-0.36cm}
\caption{Qualitative results of our model in 1-way 1-shot segmentation on PASCAL-5$^i$. The figure is best viewed in color with 300\% zooming-in.}
\vspace{0.36cm}
\label{fig:4}
\end{figure*}

\begin{table}[t]
\renewcommand\arraystretch{1.1}
\setlength\tabcolsep{3.6pt}
   \centering
   \small
   \caption{$mean{\text -}IoU$ of 1-shot and 5-shot segmentation on PASCAL-5$^i$. Our full approach outperforms all other models with the same backbone of ResNet-50. The best results among ResNet-50 based models are in bold.}
      \begin{tabular}{c|c|ccccc} 
\toprule
Backbone & Methods &split-1 & split-2 & split-3 & split-4 & mean\\ 
\hline
\multicolumn{7}{c}{Mean-IoU (1-shot)} \\ 
\hline
\multirow{3}{*}{VGG 16}    & SG-one~\cite{SG-one} & 40.2 & 58.4 & 48.4 & 38.4 & 46.3\\
                          & PANet~\cite{PANet} & 42.3 & 58.0 & 51.1 & 41.2 & 48.1 \\
                          & FWP~\cite{FWP} & 47.0 & 59.6 & 52.6 & 48.3 & 51.9 \\ 
                          & REF~\cite{zhang2021rich}& 46.5& 60.2& 52.1& 44.4& 50.8\\
\hline
\multirow{6}{*}{ResNet-50}    & CANet~\cite{CANet} & 52.5 & 65.9 & 51.3 & 51.9 & 55.4 \\
                          & PGNet~\cite{PGNet} & 56.0 & 66.9 & 50.6 & 50.4 & 56.0\\
                          & CRNet~\cite{CRNet} & - & - & - & - & 55.7\\
                          & PPNet~\cite{liu2020part} & 47.8 & 58.8 & 53.8 & 45.6 & 51.5\\
                          & PMMs~\cite{yang2020prototype} & 55.2 & 66.9 & 52.6 & 50.7 & 56.3 \\
                          & PFENet~\cite{tian2020pfenet} & 61.7 & \textbf{69.5} & 55.4 & 56.3 & 60.8\\
\cline{2-7}
                          & \textbf{SRPNet} (ours)& \textbf{62.8}\tiny{$\pm$0.16} & 69.3\tiny{$\pm$0.12} & \textbf{55.8}\tiny{$\pm$0.15} & \textbf{58.1}\tiny{$\pm$0.13} & \textbf{61.5}\tiny{$\pm$0.14} \\
\hline
\hline
\multicolumn{7}{c}{Mean-IoU (5-shot)} \\ 
\hline
\multirow{3}{*}{VGG 16}    & SG-one~\cite{SG-one} & 41.9 & 58.6 & 48.6 & 39.4 & 47.1\\
                          & PANet~\cite{PANet} & 51.8 & 64.6 & 59.8 & 46.5 & 55.7\\
                          & FWP~\cite{FWP} & 50.9 & 62.9 & 56.5 & 50.0 & 55.1\\ 
                          & REF~\cite{zhang2021rich} & 48.8 &  61.4 & 52.8& 46.2& 52.3\\
\hline
\multirow{6}{*}{ResNet-50}    & CANet~\cite{CANet} & 55.5 & 67.8 & 51.9 & 53.2 & 57.1\\
                          & PGNet~\cite{PGNet} & 57.7 & 68.7 & 52.9 & 54.6 & 58.5\\
                          & CRNet~\cite{CRNet} & - & - & - & - & 55.8 \\
                          & PPNet~\cite{liu2020part} & 58.4 & 67.8 & 64.9 & 56.7 & 62.0\\
                          & PMMs~\cite{yang2020prototype} & 56.3 & 67.3 & 54.5 & 51.0 & 57.3\\
                          & PFENet~\cite{tian2020pfenet} & 63.1 & \textbf{70.7} & \textbf{55.8} & 57.9 & 61.9\\
\cline{2-7}
                          & \textbf{SRPNet} (ours)& \textbf{64.3}\tiny{$\pm$0.10} & 70.3\tiny{$\pm$0.12} & 55.1\tiny{$\pm$0.13} & \textbf{60.5}\tiny{$\pm$0.09} & \textbf{62.6}\tiny{$\pm$0.11}\\
\bottomrule
\end{tabular}
   \label{tab:2}

\end{table}

\begin{table}[t]
\renewcommand\arraystretch{1.1}
\setlength\tabcolsep{3.6pt}
   \centering
   \small
   \caption{\RV{Dice of 1-shot and 5-shot segmentation on PASCAL-5$^i$.}}
      \begin{tabular}{c|c|ccccc} 
\toprule
Backbone & Methods &split-1 & split-2 & split-3 & split-4 & mean\\ 
\hline
\multicolumn{7}{c}{Dice (1-shot)} \\ 
\hline
\multirow{2}{*}{VGG 16}    & SG-one~\cite{SG-one} & 57.3 & 73.7 & 65.2 & 55.5 & 62.9\\
                          & PANet~\cite{PANet} & 59.5 & 73.4 & 67.6 & 58.4 & 64.7 \\
\hline
\multirow{5}{*}{ResNet-50}    & CANet~\cite{CANet} & 68.9 & 79.4 & 67.8 & 68.3 & 71.2 \\
                          & PPNet~\cite{liu2020part} & 64.7 & 74.1 & 70.0 & 62.6 & 67.9\\
                          & PMMs~\cite{yang2020prototype} & 71.1 & 80.2 & 68.9 & 67.3 & 71.9 \\
                          & PFENet~\cite{tian2020pfenet} & 76.3 & \textbf{82.0} & 71.3 & 72.0 & 75.4\\
\cline{2-7}
                          & \textbf{SRPNet} (ours)& \textbf{77.2}\tiny{$\pm$0.24} & 81.9\tiny{$\pm$0.20} & \textbf{71.6}\tiny{$\pm$0.21} & \textbf{73.5}\tiny{$\pm$0.24} & \textbf{76.2}\tiny{$\pm$0.24} \\
\hline
\hline
\multicolumn{7}{c}{Dice (5-shot)} \\ 
\hline
\multirow{2}{*}{VGG 16}    & SG-one~\cite{SG-one} & 59.1 & 73.9 & 65.4 & 56.5 & 63.7\\
                          & PANet~\cite{PANet} & 68.2 & 78.5 & 74.8 & 63.5 & 71.3\\
\hline
\multirow{5}{*}{ResNet-50}    & CANet~\cite{CANet} & 71.4 & 80.8 & 68.3 & 69.5 & 72.5\\
                          & PPNet~\cite{liu2020part} & 73.7 & 80.8 & 78.7 & 72.4 & 76.4\\
                          & PMMs~\cite{yang2020prototype} & 72.0 & 80.5 & 70.6 & 67.5 & 72.7\\
                          & PFENet~\cite{tian2020pfenet} & 77.4 & \textbf{82.8} & \textbf{71.6} & 73.3 & 76.2\\
\cline{2-7}
                          & \textbf{SRPNet} (ours)& \textbf{78.3}\tiny{$\pm$0.12} & 82.6\tiny{$\pm$0.16} & 71.1\tiny{$\pm$0.18} & \textbf{75.4}\tiny{$\pm$0.10} & \textbf{77.0}\tiny{$\pm$0.15}\\
\bottomrule
\end{tabular}
   \label{tab:dice}

\end{table}%

\subsection{Results on PASCAL-5$^i$}

{We compare our model with previous few-shot segmentation methods under the metric of $mean{\text -}IoU$. The performances of 1-shot and 5-shot segmentation are shown respectively in \tablename~\ref{tab:2}. Our proposed SRPNet outperforms the state-of-the-art methods, \eg, PFENet~\cite{tian2020pfenet}. We report standard deviations with respect to the random seed after running experiments multiple times. During testing, we average the results from 5 runs with different random seeds (thus different testing images), each run containing 1,000 episodes. The very small standard deviation values demonstrate the robustness of our proposed model.} 

{We outperforms PFENet~\cite{tian2020pfenet} by 0.7\%, which is statistically significant under the standard deviation of 0.15. However, PFENet uses prior information from the ImageNet pre-trained model and employs additional context modules, \ie, Pyramid Pooling Module (PPM)~\cite{PSPNet}, to enhance their high-level feature modeling. More importantly, we adopt a different strategy from PFENet, \ie, we use the prototypical method while PFENet uses the parametric method (see Line 16-24). The prototypical way is more lightweight than the parametric way, but parametric methods achieve better performance than prototypical methods recently. We analyze the weaknesses of the prototypical way and propose SRPNet to address these issues, resulting in significant performance gain compared to our prototypical baseline and better results than PFENet, the state-of-the-art parametric method.}

To better evaluate the performance of our proposed approach, we use Dice as another evaluation metric to report our results in \tableautorefname~\ref{tab:dice}. It is difficult to segment objects by a few support images. Few-shot segmentation is challenging while valuable for practical applications. {Our proposed approach has achieved 61.5\% mean-IoU and 76.2\% Dice under the most challenging 1-shot segmentation setting, which demonstrates the effectiveness of the proposed SRPNet.}

\begin{figure*}[t]
\begin{center}
\includegraphics[width =\textwidth]{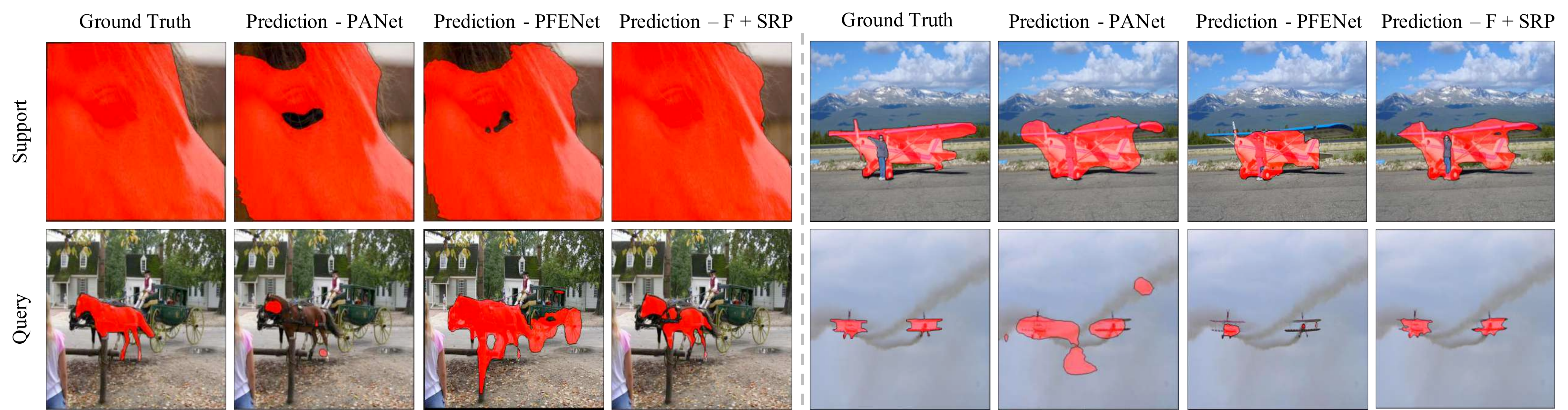}
\caption{Comparison of our SRPNet to PANet~\cite{PANet} \& PFENet~\cite{tian2020pfenet} on qualitative results in 1-way 1-shot segmentation on PASCAL-5$^i$. The figure is best viewed in color with 300\% zooming-in.}
\vspace{0.16cm}
\label{fig:5}
\end{center}
\end{figure*}

Qualitative results of 1-way 1-shot segmentation on PASCAL-5$^i$ are shown in \figurename~\ref{fig:4}. None of the shown case is trivial due to reasons like the presence of multiple objects (\eg, bicycles, cow$^1$ and motorbike$^2$), the huge variances in object size (\eg, bird$^1$, aeroplane$^1$ and cow$^2$) and appearance (\eg, bird$^2$), different viewing angles (motorbike$^1$ and horses), as well as the requirements of retaining object details. Using only a single annotated supporting image, satisfactory segmentation results can be obtained on unseen classes, which demonstrates the powerful learning and generalization capabilities.
{Our prototype regularization - F+SRP - significantly and intuitively improves the expression of details when applying extracted prototypes onto the support images themselves, as shown in \figurename~\ref{fig:5}. For example, in the segmentation of horse, PANet and PFENet get confused by the eye of the horse when generating prototype, resulting in a disordered prediction on the query image. Whereas, given that our method benefits from supervision of prototype generation using support ground-truth, it obtains relatively complete predictions, retaining details such as horse legs and tails. The advantages of prototype supervision are also observed when transferring knowledge between objects of different sizes.} 
In the case of airplane, the support image contains a normal-sized airplane, while the query image is two tiny airplanes. PANet and PFENet fail in identifying the details and depicts roughly the overall contours of the supporting aeroplane, which results in only blurry encircling objects when predicting the query. Some of the circles are even meaningless, perhaps due to the failure to effectively identify and eliminate human objects in the supporting image. Our approach, as opposed, narrowed down the prediction to a more precise range, restoring the details of the tiny aeroplane, including the wings. 

\begin{figure}[t]
\begin{center}
\includegraphics[width =0.86\textwidth]{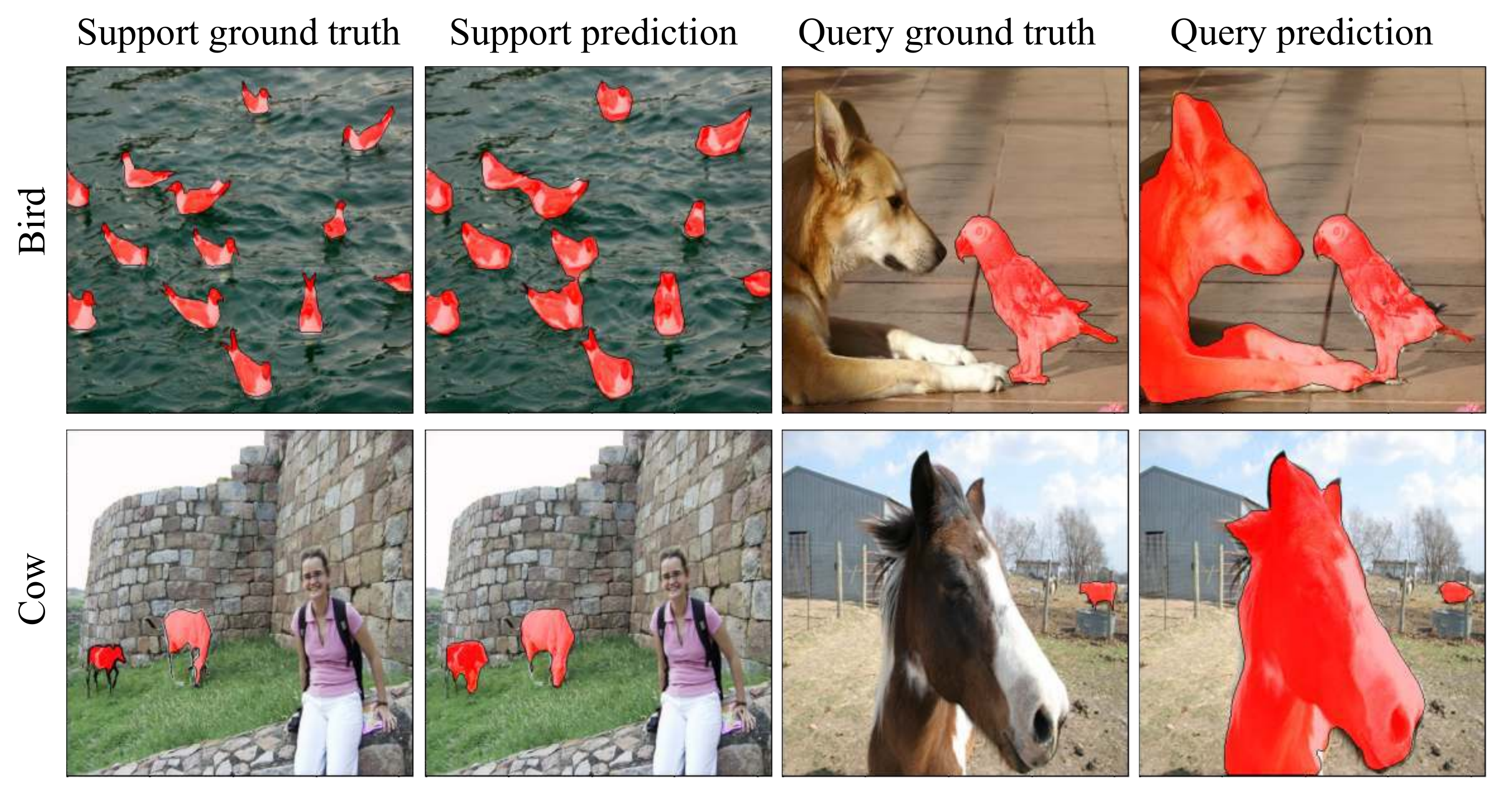}
\caption{Exemplary failure cases in 1-way 1-shot segmentation. {Two types of errors are listed. We speculate that one is caused by the model’s simple interpretation of the segmentation task as removing the background or retaining all targets except for the monotonous background (\eg, removing the lake in support as well as the ground in query, in the bird case), and the other is caused by the similarity of objects (misidentifying horses as cows, in the cow case)}}
\label{fig:6}
\end{center}
\end{figure}

We still fail some challenging cases as shown in \figurename~\ref{fig:6}. Two types of errors are listed. We speculate that one is caused by the model's simple interpretation of the segmentation task as removing the background or retaining all targets except for the monotonous background (\eg, removing the lake in support as well as the ground in query, in the bird case), and the other is caused by the similarity of objects (misidentifying horses as cows, in the cow case). 

\subsection{Results on MS COCO}

We test our full approach F+SRP+IQI on MS COCO, which contains 80 object categories. Experimentally, a subset of the original large-scale is evaluated, which contain 39,107 (40 classes) training samples, 5,895 validation samples (20 classes) and 9,763 testing samples (20 classes). Our SRP achieves $mean{\text -}IoU$ improvement of 2.7\% in the 1-shot setting and 6.0\% in 5-shot setting, compared to PFENet and as shown in \tablename~\ref{tab:3}. 

{MS COCO has more classes than PASCAL-5$^i$, resulting in more confusion among classes. A better prototype that contains discriminative clues against visual-similar classes helps enhance the segmentation performance. Our proposed Self-Regularized Prototypical Networks optimizes the prototype generation and collects a set of iteratively refined prototypes, producing more representative and generalizable prototypes and better supporting segmentation on MS COCO. Additionally, there are more training samples in MS COCO, which also contributes to the optimization of prototype generation via self-regularization.}

\begin{table}
\renewcommand\arraystretch{1.1} 
\small
\centering
\caption{$mean{\text -}IoU$ of 1-shot and 5-shot segmentation on MS COCO.}
\begin{tabular}{c|c|c} 
\toprule
Method & 1-shot & 5-shot  \\ 
\hline
PANet (VGG 16)~\cite{PANet}      & 20.9 & 29.7 \\
FWP (ResNet-101)~\cite{FWP}        & 21.2 & 23.7 \\
REF (VGG 16)~\cite{zhang2021rich} &22.0 & 31.1\\
PMMs (ResNet-50)~\cite{yang2020prototype}    & 30.6 & 35.5 \\
PFENet (ResNet-101 v2)~\cite{tian2020pfenet}        & 32.4 & 37.4 \\
\hline
SRPNet (Ours, ResNet-50)       & \textbf{35.1} & \textbf{43.4}\\
\bottomrule
\end{tabular}
\label{tab:3}
\end{table}
\section{Conclusion}
\label{sec:Conclusion}
In this work, we present a self-regularized prototypical network for few-shot semantic segmentation. A direct yet effective regularization is proposed by evaluating prototypes on support set reversely, based on the observations that the generated prototype cannot consistently describe support set itself. 
Our SRPNet exploits the in-depth knowledge from the support set, and offers high-quality prototypes that can well represent each semantic category while distinguishing it from other categories.
An iterative process is adopted in query inference, which fuses multiple prototypes based on the regularized prototype to generate the final segmentation result. Besides, we adopt fidelity  as the distance metric for the first time, which achieves more evident distinctions between prototypes and feature maps. Our proposed SRPNet leads to new state-of-art performance on 1-shot and 5-shot segmentation benchmarks.

\small
\bibliography{egbib}

\end{document}